\documentclass[letterpaper, 10 pt, conference]{ieeeconf}  

\IEEEoverridecommandlockouts                              

\usepackage[pdftex]{graphicx}
\usepackage[pdftex]{color}
\usepackage{amsmath}
\usepackage{amssymb}
\usepackage{algorithm}
\usepackage{algpseudocode}
\usepackage{times}
\usepackage{float}
\usepackage{url}
\newcommand{\figref}[1]{Fig.~\ref{figure:#1}}

\title{\LARGE \bf
Remote Life Support Robot Interface System for Global Task Planning \\and Local Action Expansion Using Foundation Models
}

\author{Yoshiki Obinata$^{1}$, Haoyu Jia$^{1}$, Kento Kawaharazuka$^{1}$, Naoaki Kanazawa$^{1}$ and Kei Okada$^{1}$
\thanks{$^{1}$The authors are with the Department of Mechano-Informatics, Graduate School of Information Science and Technology, The University of Tokyo, 7-3-1 Hongo, Bunkyo-ku, Tokyo, 113-8656, Japan. [obinata, jia, kawaharazuka, kanazawa, k-okada]@jsk.imi.i.u-tokyo.ac.jp
}
}

\begin{document}

\maketitle
\thispagestyle{empty}
\pagestyle{empty}

\begin{abstract}
Robot systems capable of executing tasks based on language instructions have been actively researched. It is challenging to convey uncertain information that can only be determined on-site with a single language instruction to the robot. In this study, we propose a system that includes ambiguous parts as template variables in language instructions to communicate the information to be collected and the options to be presented to the robot for predictable uncertain events. This study implements prompt generation for each robot action function based on template variables to collect information, and a feedback system for presenting and selecting options based on template variables for user-to-robot communication. The effectiveness of the proposed system was demonstrated through its application to real-life support tasks performed by the robot.
\end{abstract}

\section{Introduction}

Robot systems for life support have been researched for a long time \cite{6218159}. In the future, it will be important for people to be able to entrust high-degree-of-freedom life support robots with tasks at remote locations to improve their quality of life. People should have more free time while the robot is performing tasks, and the time people need to intervene in the robot's operation should be as short as possible.

Accurately telling tasks to robots has long been a challenge. However, recent advances in Large Language Models (LLMs) are revolutionizing this process, enabling robots to interpret instructions communicated in natural language and convert them into sequences of action functions. Yet, a system that relies solely on natural language cannot effectively handle tasks that require the robot to query the user based on local information.

\begin{figure}[tb]
  \centering
  \includegraphics[width=\columnwidth]{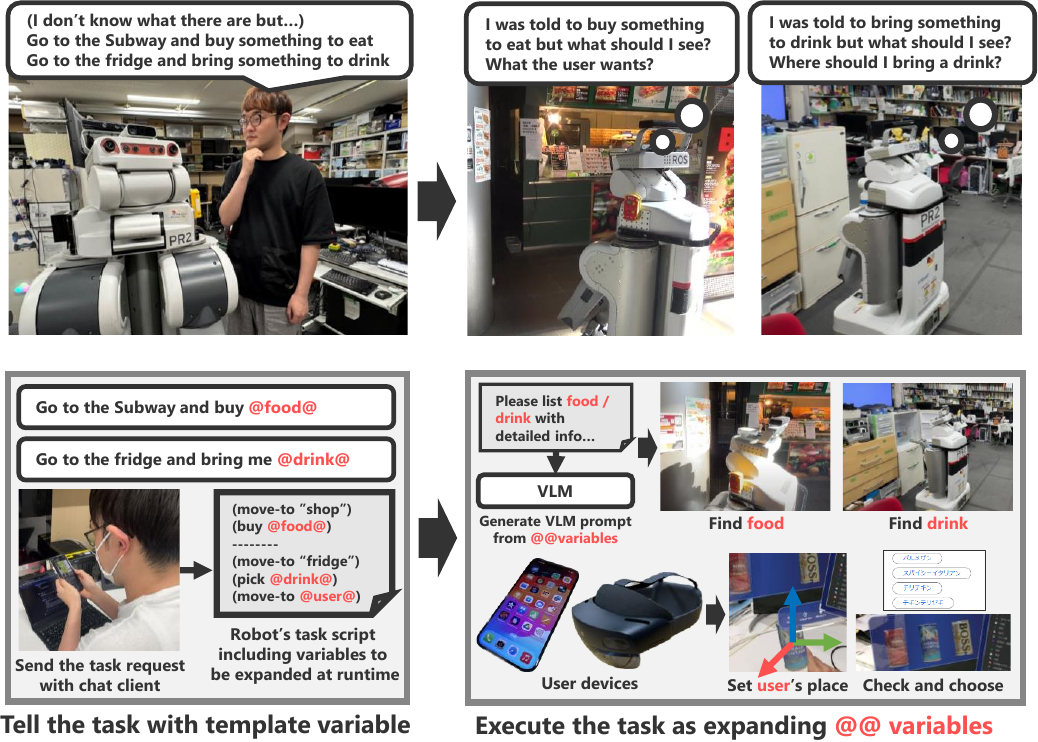}
  \caption{Problem solved in this research (top) and the proposed system (bottom). The user requests the robot to buy food and bring drinks, but the robot does not know what to look for on-site, what the user wants, or where to deliver the items. In this research, we propose using natural language instructions with template variables surrounded by @ to specify the information that needs to be collected and the queries that need to be made when the user instructs the robot. When executing action functions containing template variables, the robot sets up recognizers (Vision Language Model prompts) according to the content of the template variables, presents options to the user's operation device, and continues task actions by receiving feedback from these options and the operation UI.}
  \label{figure:concept}
\end{figure}

\figref{concept} shows an example of tasks requiring local information collection and user queries and our proposed solution in this research. The user requests the robot to buy food at a food shop and fetch drinks from a refrigerator. However, it is difficult to convey complete information about the tasks beforehand because the user does not know what food or drinks are available on site. When the robot arrives at the food shop or refrigerator, it needs to collect the data necessary for task continuation and find out what the user wants and where to deliver it.

In this research, we propose a novel concept of generating robot task scripts with template variables. This approach not only resolves geometric uncertainties but also presents the user with options based on on-site information, and executes the generated script's template variables. The user communicates with the robot in natural language through a text chat, including task template variables, and the robot then generates a task script composed of its action functions. The template variables are described in the task script, and the robot sets up recognizers or presents options to the user to expand these variables. In this research, we have used tap operations on smartphones and operations with AR glasses to ensure intuitive user operation.

\section{Related Works}
\begin{figure*}[tb]
  \centering
  \includegraphics[width=0.85\linewidth]{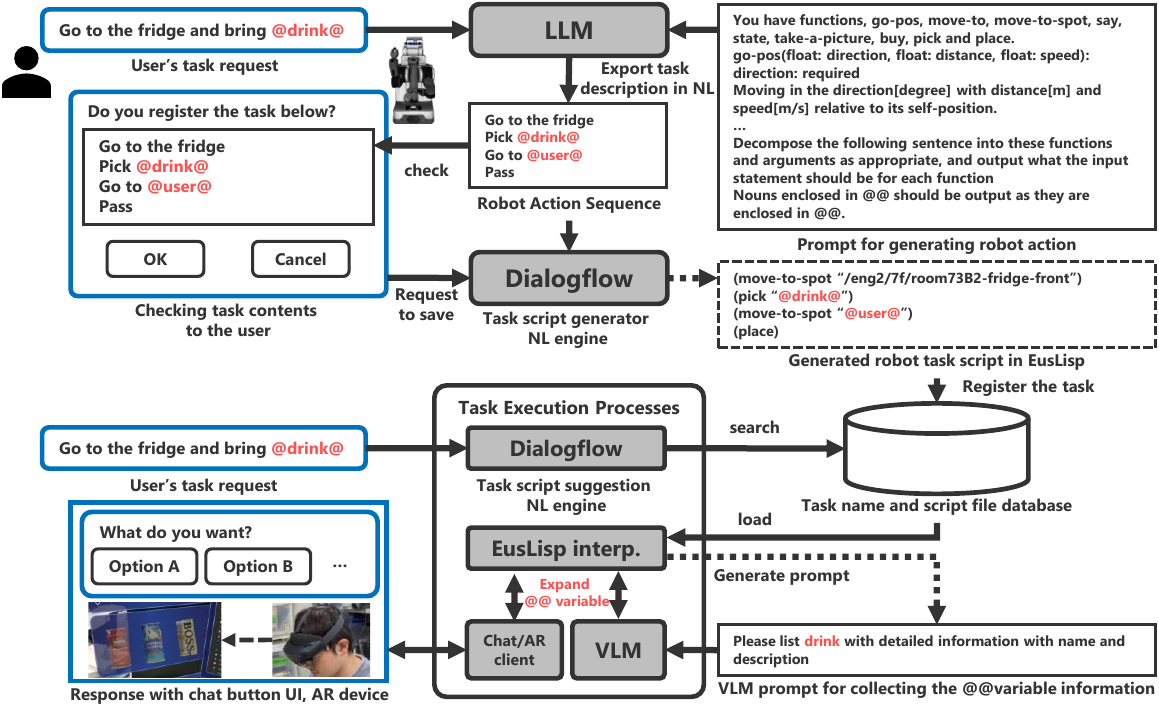}
  \caption{Overview of the proposed system in this research. The system has two stages: the stage where the user communicates tasks to the robot, generates and saves task scripts, and the stage where task scripts are executed. In the stage of generating Task Scripts, the user communicates task instructions with template variables surrounded by @ to the robot in text. The robot inputs the instructions into the LLM along with prompts about the robot's functions and presents the decomposed robot action sequence to the user for confirmation. Once the user approves, the action sequence is input into Dialogflow for conversion into an EusLisp script, which is then registered in the database. In the stage of executing Task Scripts, when the user communicates a task instruction to the robot, the robot searches for similar tasks registered in the past and loads the script. The robot then executes the EusLisp script to perform the task. If the robot encounters template variables surrounded by @ during task execution, it generates appropriate VLM prompts to obtain information from the VLM or queries the user via chat buttons or AR devices to continue the operation.}
  \label{figure:whole-system}
\end{figure*}
There is research where users instruct mobile manipulators through natural language. SayCan \cite{saycan} is a pioneering effort in this area, and research on generating robot action sequences from language has been done \cite{10160591,HuangXXCLFZTMCS22}. Recently, this approach has been widely used in life support robot competitions \cite{obinata2023foundation,shirasaka2023self}. There is also research on robots querying humans in uncertain situations during task execution \cite{ren2023robots}.

Intuitive interfaces for users to instruct robots are important for the spread of life support robots. There are studies on interfaces for intuitively operating high-degree-of-freedom robots. For example, research on users instructing dual-arm robots to manipulate objects using tangible cubes \cite{doi:10.1080/01691864.2023.2239316}, research on users instructing robots to operate home appliances using tablets \cite{azuma2012instruction}, and research on instructing navigation for mobile robots in real space using AR \cite{iglesius2024mrnab}.

In this research, we adopt natural language as an intuitive and expressive interface for conveying tasks from users to robots. We propose a system that allows robots to collect and present information on-site that the user cannot convey to the robot due to a lack of prior knowledge before task execution. To enable users to intuitively instruct robots with geometric information that is difficult to convey verbally, we use AR glasses to present information from users to robots in real space. This research proposes a concept of a system for transmitting information between robots and humans using natural language, touch UI, and AR glasses UI as needed.

\section{Methods}
\figref{whole-system} shows an overview of the proposed system in this research. We propose a system that uses template variables surrounded by @ on both the robot and user interface sides to explicitly specify uncertain information during the instruction stage, allowing the robot to expand it on-site.

\subsection{Generation of Task Scripts with Template Variables in Natural Language}\label{generate}
We explain the method by which the user instruct the robot on tasks.

First, the user instructs the robot in natural language. At this time, the user can introduce template expressions in the instruction text to specify information that is not determined at the instruction stage, which the robot will collect and present to the user during task execution. In the example in \figref{whole-system}, the user requests the robot to go to the fridge and bring a drink. Since the user does not know what drinks are in the fridge at the instruction stage, they use the template variable surrounded by @ as in ``Go to the fridge and bring @drink@.''

When the robot receives the instruction, it inputs the user's instruction text and pre-set prompts to the LLM to generate the robot's action sequence. The prompts include descriptions of the robot's pre-existing action functions, instructions to convert the input instruction text into a sequence of functions, and instructions to keep variables surrounded by @ as they are. The robot receives the robot action sequence from the LLM and confirms its correctness with the user. When the user approves, the robot generates a script in EusLisp \cite{matsui1990euslisp} executable by the robot from the robot action sequence.

To generate EusLisp scripts executable by the robot from the robot action sequence written in natural language without hallucinations, we use Google's natural language processing engine Dialogflow \cite{dialogflow} as the task script generator. In the Dialogflow of the task script generator, we have trained a model that outputs the name of the robot's action function and its arguments from a short natural language sentence in our previous research \cite{10.1007/978-3-031-44981-9_31}. The robot inputs each robot action into this model, obtains the corresponding EusLisp function name and arguments, and generates an EusLisp script executable by the robot.

When generating EusLisp functions with the Dialogflow of the task script generator, it may be necessary to resolve navigation symbol names. For example, in \figref{whole-system}, when converting the action ``Go to the fridge'' into an EusLisp function, it is necessary to set the location of ``the fridge.'' The Dialogflow of the task script generator has registered the names and aliases of location symbols in the robot's map. During the generation of EusLisp functions, it changes the function arguments to the names of location symbols registered in the map, enabling navigation using the system from previous research \cite{6224965}.

Once the task script is generated from the user's instructions, the user's instruction text and task script are registered in the database. When the user later communicates similar instructions to the robot, the robot outputs similar tasks from the past instructions using Dialogflow, retrieves the task from the database, and executes the task immediately without generating the task script.

This research uses GPT-4 \cite{NEURIPS2020_1457c0d6, achiam2023gpt} for the LLM.

\subsection{Information Collection for Template Variables During Task Execution}\label{collect}
\begin{figure}[tb]
  \centering
  \includegraphics[width=\columnwidth]{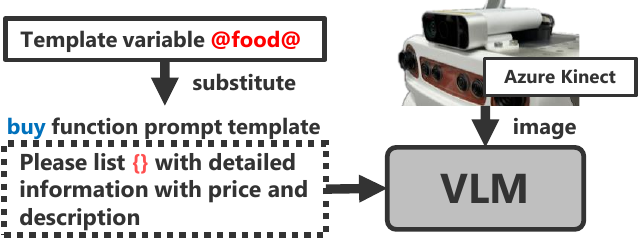}
  \caption{System for the robot to collect information about arguments surrounded by templates on-site. Prompt templates for input to the VLM corresponding to the functions are prepared in advance, and the user's input is substituted to create the VLM's language input. The image input to the VLM is provided by the high-resolution camera Azure Kinect mounted on the robot.}
  \label{figure:vlm-extract-info}
\end{figure}

We explain the method by which the robot collects information about template variables during task execution.

While executing the task script generated in Sec. \ref{generate}, the robot expands template arguments surrounded by @. To expand these arguments, the robot needs to collect on-site information to query the user. For example, as shown in \figref{vlm-extract-info}, for the buy function, the robot needs to collect information about what is being sold and at what price. In this research, we use Vision Language Model (VLM) for on-site information collection based on template variables, inputting the current image seen by the robot as the image input and dynamically generated prompts from the function and argument name as the language input. There is a prompt template for each function, and the prompt is completed by substituting the template arguments surrounded by @.

In this research, we implement a system to collect information about the template arguments for the buy function and the pick function, using the prompt shown in \figref{vlm-extract-info} for the buy function and ``Please list \{\} with detailed information with name and description'' for the pick function.

To format the output when sending the VLM's output to the user's device, we input the VLM's output to the LLM (OpenAI's GPT-4 in JSON mode) and output the result in JSON format. We use Microsoft's Azure Kinect, mounted on the robot's head, as the camera for input to the VLM.

\subsection{User Interface for Expanding Template Variables at Runtime}
We explain the interface system for the robot to instruct the user to expand template variables.

We use a chat system \cite{10.1007/978-3-031-44981-9_31} and an AR interface as the interfaces for the robot to prompt the user to select actions for expanding template variables. The chat interface is used for the robot to present option buttons, and the AR interface is used to present options and instruct the robot on the pose of objects around the user.

\begin{figure}[tb]
  \centering
  \includegraphics[width=0.6\columnwidth]{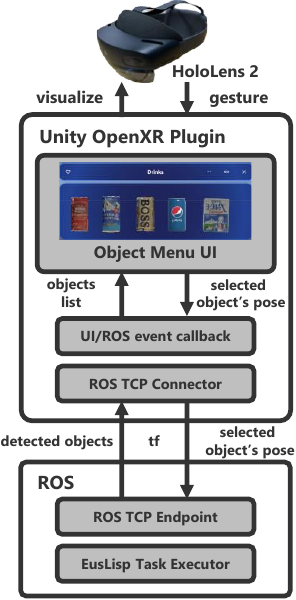}
  \caption{AR system for presenting template variable options from the robot to the user and presenting object poses from the user to the robot. The AR glasses menu UI displays a list of objects detected by the robot based on the robot's information. When the user selects an object from the menu UI and places it around themselves, as shown in \figref{ar-ui}, the pose is sent to the robot. The AR device uses Microsoft HoloLens 2, the runtime uses Unity OpenXR Plugin, the UI is created with Microsoft's Mixed Reality Toolkit 2, and communication between the AR device and ROS uses ROS TCP Connector / Endpoint.}
  \label{figure:ar-system}
\end{figure}

\begin{figure}[tb]
  \centering
  \includegraphics[width=\columnwidth]{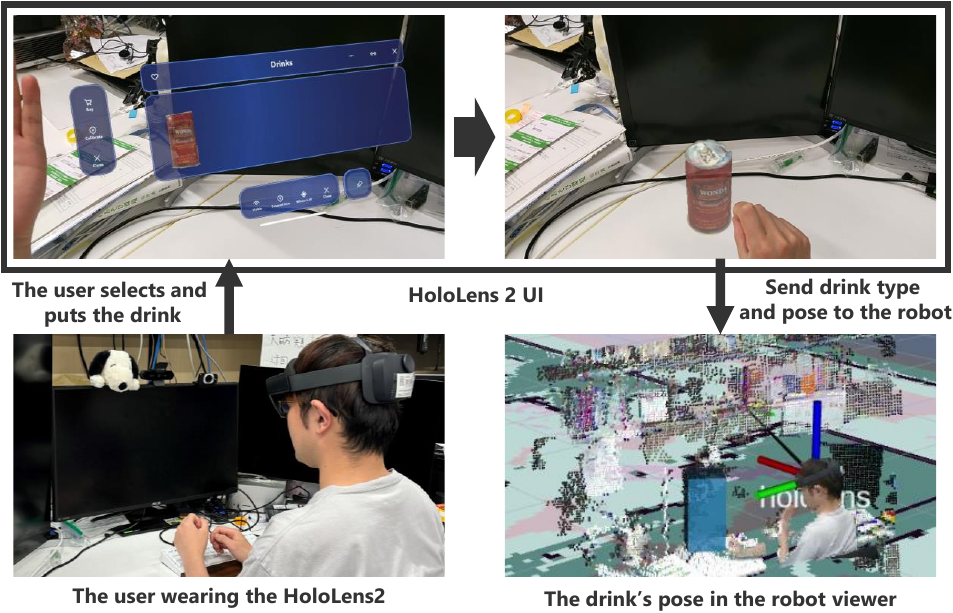}
  \caption{The user manipulates objects in AR, and the information is transmitted to the robot. When the user places a drink from the menu on the table, the type and pose of the drink are transmitted to the robot.}
  \label{figure:ar-ui}
\end{figure}

\figref{ar-system} shows the system for the user to expand template variables using the AR interface. The AR glasses menu UI displays a list of objects detected by the robot. As shown in \figref{ar-ui}, when the user grabs an object from the menu UI and places it around themselves in the AR screen, its pose is sent to the robot. Based on this pose, the robot expands the template variables, obtains the necessary object location information, and continues the operation. The AR device uses Microsoft HoloLens 2, the runtime uses Unity OpenXR Plugin, the UI is created with Microsoft's Mixed Reality Toolkit 2, and communication between the AR device and the robot's middleware ROS \cite{quigley2009ros} uses Unity ROS TCP Connector / Endpoint.

\begin{figure}[tb]
  \centering
  \includegraphics[width=\columnwidth]{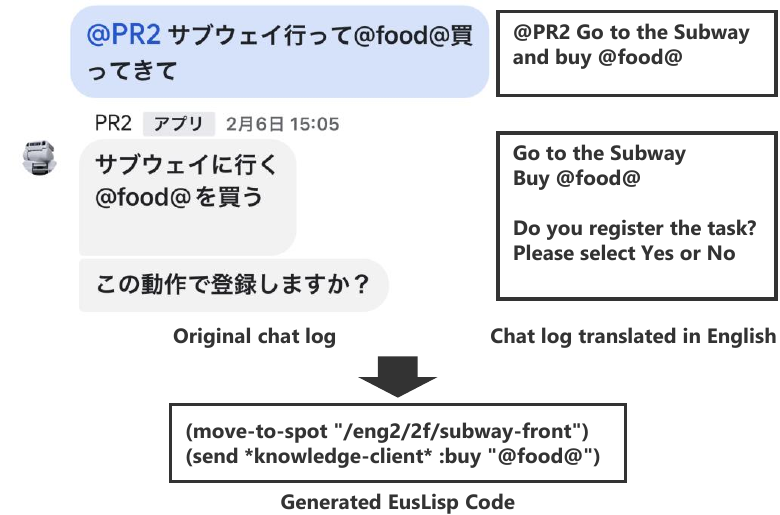}
  \caption{Chat log and generated code when the user requested to buy @food@ at a sandwich shop. Note that ``*knowledge-client*'' included in the generated code is an instance defined in the system's implementation, and attention should be focused on the method functions only.}
  \label{figure:subway-gen}
\end{figure}
\begin{figure*}[tb]
  \centering
  \includegraphics[width=\linewidth]{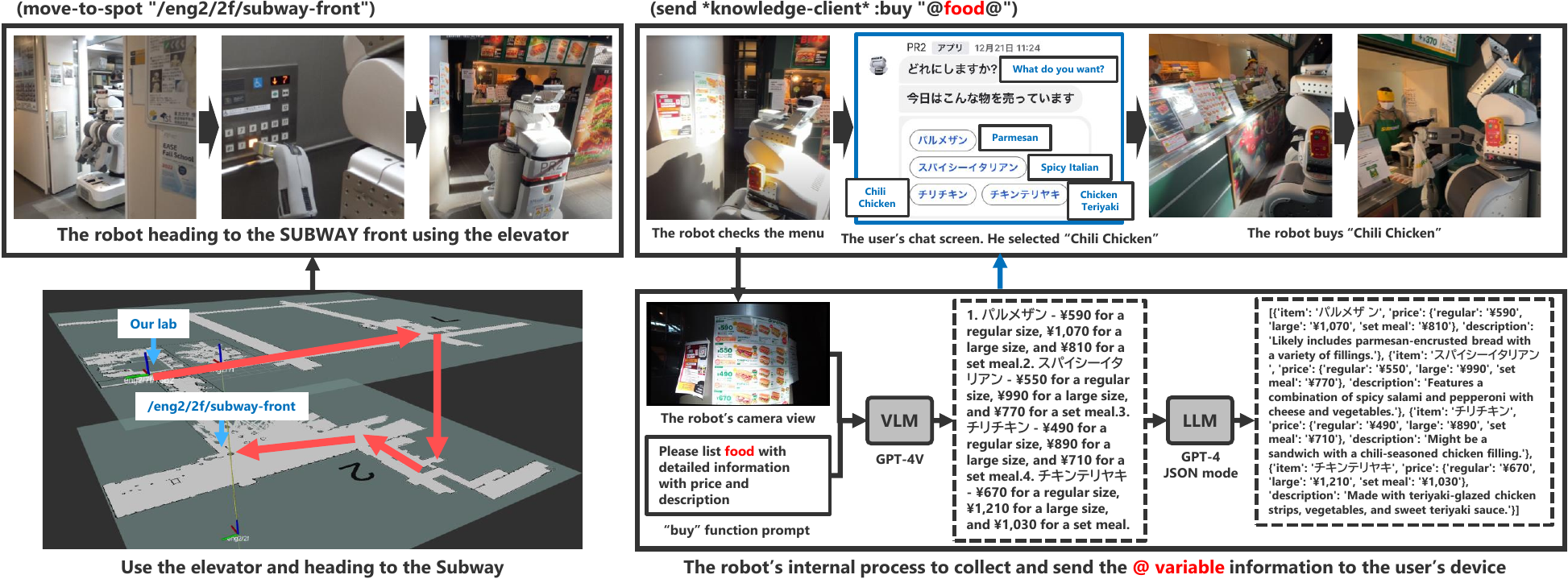}
  \caption{The result of executing the task of going to Subway and buying @food@. The robot used an elevator to reach the floor where the map symbol for Subway was located. Upon arriving at Subway, the robot collected information about @food@ by capturing an image of the menu and inputting it to the VLM along with the prompt for the buy function. The VLM's output was converted to JSON format by the LLM and presented to the user in the chat interface with options ``Parmesan,'' ``Spicy Italian,'' ``Chili Chicken,'' and ``Chicken Teriyaki.'' The user selected ``Chili Chicken'' from the chat interface. The robot told the staff that it wanted to buy ``Chili Chicken,'' received the item, and completed the purchase. The Japanese characters in the VLM output text represent 1. ``Parmesan,'' 2. ``Spicy Italian,'' 3. ``Chili Chicken,'' and 4. ``Chicken Teriyaki.''}
  \label{figure:exp-subway}
\end{figure*}
\begin{figure}[tb]
  \centering
  \includegraphics[width=\columnwidth]{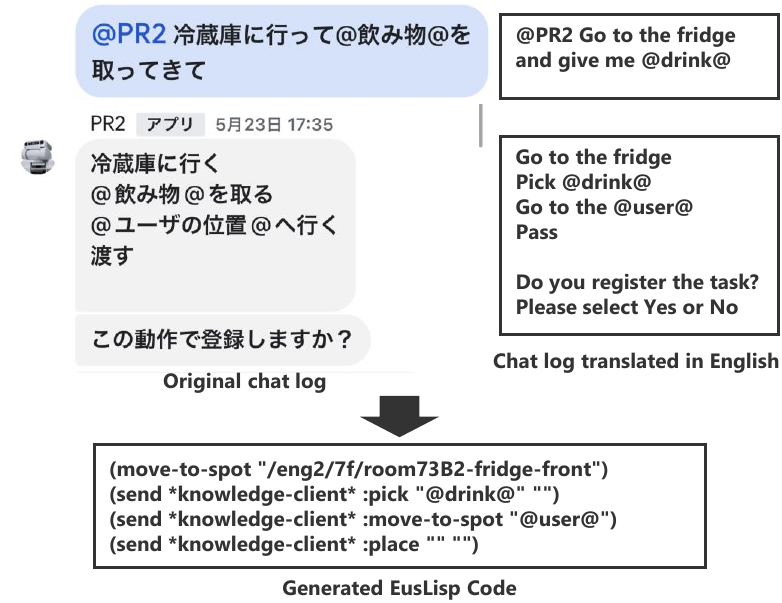}
  \caption{Chat log and generated code when the user requested to get @drink@ from the fridge. Note that ``*knowledge-client*'' included in the generated code is an instance defined in the system's implementation, and attention should be focused on the method functions only.}
  \label{figure:fridge-gen}
\end{figure}
\begin{figure*}[tb]
  \centering
  \includegraphics[width=\linewidth]{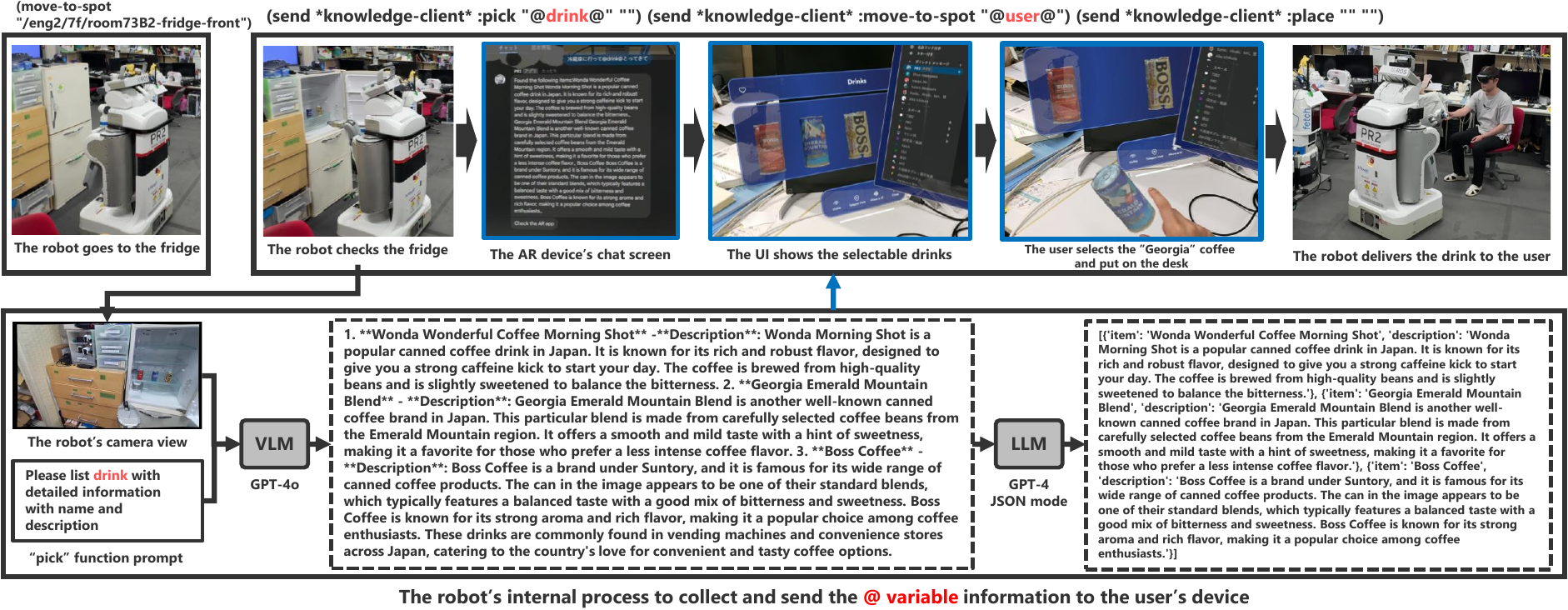}
  \caption{The result of executing the task of going to the fridge and getting @drink@. The robot moved to the fridge. Upon arriving at the fridge, the robot collected information about @drink@ by opening the fridge, capturing an image of the contents, and inputting it to the VLM along with the prompt for the pick function. The VLM's output was converted to JSON format by the LLM and sent to the user's AR device chat interface with information about the drinks. Simultaneously, the robot sent a list of drinks in the fridge to the user's AR device, displaying the available drinks to the user. The user selected the drink ``Georgia'' and placed it on the table in the AR interface. The robot then picked ``Georgia'' from the fridge, moved to the user's table, and handed ``Georgia'' to the user.}
  \label{figure:exp-fridge}
\end{figure*}

\section{Experiments}
To demonstrate the usefulness of the proposed system, we applied it to real-world robot tasks and conducted experiments. All experiments were conducted in the Engineering Building 2 at The University of Tokyo, using the dual-arm mobile humanoid PR2 \cite{bohren2011towards}.

\subsection{Purchasing Behavior Requiring Expansion of Purchase Items at a Sandwich Shop}\label{sec:exp-subway}
As shown in \figref{subway-gen}, the user sent a message to the robot saying, "Go to the Subway and buy @food@." The robot responded with an action sequence: 1. "Go to the Subway", 2. "Buy @food@", and generated the corresponding EusLisp code to execute these actions. For "Go to the Subway," the location of the Subway was converted to the map symbol ``/eng2/2f/subway-front'' in the robot's map.

The robot navigated to ``/eng2/2f/subway-front'' at first. Since this map symbol was on the 2nd floor of the building while the robot was on the 7th floor, it used the elevator to reach the destination. Upon arriving in front of Subway, the robot pointed its camera at the menu to collect information about the food surrounded by @ and input the camera image to the VLM along with the prompt for the buy function. The VLM outputs the names and prices of the menu items it reads. The LLM converted the VLM's output to JSON format and presented the options ``Parmesan,'' ``Spicy Italian,'' ``Chili Chicken,'' and ``Chicken Teriyaki'' to the user in the chat interface. The user selected ``Chili Chicken'' from the chat interface. The robot told the staff that it wanted to buy ``Chili Chicken,'' received the item, and completed the task. In this experiment, @food@ was expanded to ``Chili Chicken.''

Comparing the VLM's output with the actual menu image information, ``Chili Chicken'' and ``Chicken Teriyaki'' were present on the actual menu, but ``Parmesan'' and ``Spicy Italian'' were not. Among the prices, only the price for ``Chili Chicken'' was correct.

In this experiment, the VLM model used was GPT-4V, and the instruction for the LLM to convert the VLM's output to JSON was: ``We asked the vision model the question '\$question' and got the answer '\$answer'. Please summarize and list the results with JSON format. For each item, please write the name of the item in the 'item' key, price of the item in the 'price' key, and detailed information about the item in the 'description' key. Please do not add any other information.'' The \$question was the prompt for the buy function input to the VLM, and the \$answer was its output.

\subsection{Delivery Behavior Requiring Expansion of Pick Items and User Location at a Fridge}
As shown in \figref{fridge-gen}, the user sent a message to the robot saying, "Go to the fridge and give me @drink@." The robot responded with an action sequence: 1. "Go to the fridge", 2. "Pick @drink@", 3. "Go to the @user@", 4. "Pass", and generated the corresponding EusLisp code to execute these actions. For "Go to the fridge," the fridge's location was converted to the map symbol ``/eng2/7f/room73B2-fridge-front'' in the robot's map.

The robot navigated to ``/eng2/7f/room73B2-fridge-front'' at first. Upon arriving in front of the fridge, the robot opened the fridge door and pointed its camera inside to collect information about the drink surrounded by @ and input the camera image to the VLM along with the prompt for the pick function. The VLM outputs the names and descriptions of the drinks inside the fridge. The LLM converted the VLM's output to JSON format and sent the information about the drinks to the user's AR device chat interface. Simultaneously, the robot sent a list of drinks available in the fridge to the user's AR device, which displayed the available drinks to the user. The user selected the drink ``Georgia'' and placed it on the table in the AR interface. The robot then picked ``Georgia'' from the fridge, moved to the user's table, and handed ``Georgia'' to the user. In this experiment, @drink@ was expanded to ``Georgia.''

Comparing the VLM's output with the drinks available inside the fridge, Wonda Wonderful Coffee Morning Shot, Georgia Emerald Mountain Blend, and Boss Coffee were all present. The fridge was captured on the right side of the input image to the VLM, occupying a small proportion of the image, yet the VLM accurately extracted the information about the drinks.

In this experiment, the VLM model used was GPT-4o, and the instruction for the LLM to convert the VLM's output to JSON was: ``We asked the vision model the question '\$question' and got the answer '\$answer'. Please summarize and list the results with JSON format. For each item, please write the name of the item in the 'item' key and detailed information about the item in the 'description' key. If you don't know the name, please fill 'unknown' in the 'item' key. Please do not add any other information.'' The \$question was the prompt for the pick function input to the VLM, and the \$answer was its output.

\section{Discussion}
In the experiments, the user instructed the robot to collect predefined uncertain information, and the robot presented the user with multiple-choice information from the site, continuing the tasks based on user feedback. These results demonstrate that the proposed system allows the user to execute tasks according to their intentions with minimal intervention, even for tasks that require on-site information.

In the experiment described in Sec. \ref{sec:exp-subway}, the VLM model produced hallucinations, outputting information that did not exist. To avoid this, switching to a higher-performing model or adopting approaches that average outputs from multiple VLMs may be beneficial by changing the field of view multiple times or adding noise to the images.

We also discuss this system's limitations. During the buy or pick tasks, the robot navigates to a designated location to collect information necessary for expanding template variables. In real-world environments, the robot must autonomously navigate to appropriate positions and orient its camera to access the information required for template expansion. Additionally, in the buying task, it is necessary to obtain information not only from camera images but also through interactions with people to continue the task.

\section{Conclusion}
In this study, we proposed a system that introduces template expressions into task instruction sentences from users to robots, allowing users to explicitly communicate the information they want the robot to collect and present during task execution. We incorporated a VLM model for the robot to collect information and used AR glasses in addition to our previously used chat system to facilitate the transmission of geometric information between users and robots. Experiments with the real robot demonstrated that the proposed system enables users to execute tasks according to their intentions with minimal intervention, even for tasks that require on-site information.

In the future, to apply this system to many kinds of tasks, further research is needed on autonomous data collection strategies by robots for template variable expansion and on more intuitive and expressive interface systems.

\addtolength{\textheight}{-12cm}   

\bibliographystyle{junsrt}
\bibliography{main}

\end{document}